\documentclass[10pt,twocolumn,letterpaper]{article} 

\usepackage{avss}
\usepackage{times}
\usepackage{epsfig}
\usepackage{graphicx}
\usepackage{amsmath}
\usepackage{amssymb}
\usepackage{multirow}
\usepackage{float}
\usepackage{enumitem}

\usepackage{kotex}

\newcommand{\yj}[1]{{#1}}
\newcommand{\dw}[1]{{#1}}
\newcommand{\km}[1]{{#1}}

\avssfinalcopy 

\def\avssPaperID{172} 

\ifavssfinal\fi
\begin{document}

\title{CoT-Segmenter: Enhancing OOD Detection in Dense Road Scenes via Chain-of-Thought Reasoning}




\author{Jeonghyo Song$^{1}$, Kimin Yun$^{2,3}$, DaeUng Jo$^{4}$, Jinyoung Kim$^{1}$, Youngjoon Yoo$^{1,*}$\\
$^1$Department of Artificial Intelligence, Chung-Ang University, Seoul, Korea\\
$^2$Visual Intelligence Lab., ETRI, Daejeon, Korea\\
$^3$University of Science and Technology~(UST), Daejeon, Korea\\
$^4$School of Electronics Engineering, Kyungpook National University, Daegu, Korea\\
{\tt\small \{thd9592s, barraki7226\}@cau.ac.kr, kimin.yun@etri.re.kr, daeung.jo@knu.ac.kr, yjyoo3312@gmail.com}
\thanks{Corresponding author}
}

\maketitle

\begin{abstract}
Effective Out-of-Distribution (OOD) detection is critical for ensuring the reliability of semantic segmentation models, particularly in complex road environments where safety and accuracy are paramount. 
Despite recent advancements in large language models (LLMs), notably GPT-4, which significantly enhanced multimodal reasoning through Chain-of-Thought (CoT) prompting, the application of CoT-based visual reasoning for OOD semantic segmentation remains largely unexplored.
\yj{In this paper, }
through extensive analyses of the \yj{road scene anomalies}, we identify three challenging scenarios where current state-of-the-art OOD segmentation \yj{methods} consistently struggle: (1) densely packed and overlapping objects, (2) distant scenes with small objects, and (3) large foreground-dominant objects. 
To address \yj{the presented} challenges, we propose a novel CoT-based framework targeting OOD detection in road anomaly scenes.  
\yj{Our method leverages the extensive knowledge and reasoning capabilities of foundation models, such as GPT-4, to enhance OOD detection through improved image understanding and prompt-based reasoning aligned with observed problematic scene attributes. 
Extensive experiments show that our framework consistently outperforms state-of-the-art methods on \km{both standard benchmarks and our newly defined challenging subset of the RoadAnomaly dataset}, offering a robust and interpretable solution for OOD semantic segmentation in complex driving environments.}
\end{abstract}

\section{Introduction}

Reliable Out-of-Distribution (OOD) detection is essential for robust semantic segmentation, especially in safety-critical domains like autonomous driving, where failing to identify unfamiliar objects can have severe consequences. \yj{In the road scene scenario}, OOD detection aims to distinguish inputs that differ significantly from the training distribution, ensuring that autonomous systems can safely and effectively respond to unexpected or rare scenarios. Despite recent advancements, accurately identifying and segmenting OOD objects remains challenging due to real-world complexities and inherent limitations of existing models.

\begin{figure}[!t]
    \centering
    \includegraphics[width=\linewidth]{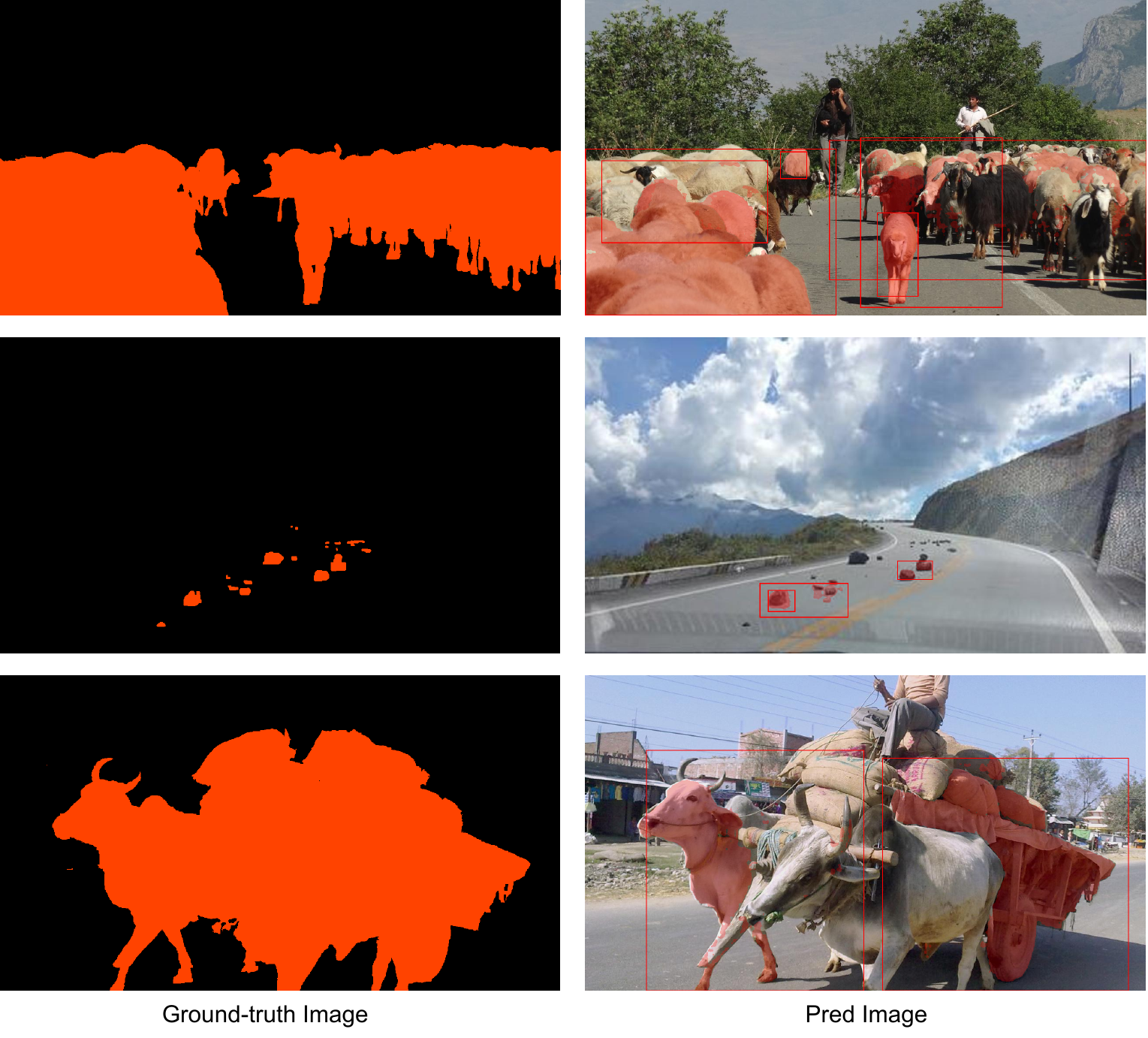}
    \caption{Visualization of three challenging scenarios affecting performance degradation in OOD segmentation models, observed in the current state-of-the-art method S2M \cite{zhao2024segment}.
    From top to bottom: (1) densely packed and overlapping objects, (2) distant scenes with small objects, and (3) large foreground-dominant objects.}
    \label{fig:bad_scene}
\end{figure}

\yj{Specifically,} we identify critical failure cases in Figure~\ref{fig:bad_scene} in which existing OOD segmentation methods consistently fail: (1) densely clustered and overlapping objects, (2) small and distant objects, and (3) scenes dominated by large foreground objects from the 
RoadAnomaly dataset \cite{lis2019detecting}, a representative benchmark for road scene understanding. 
\km{These challenging scenarios reveal a fundamental limitation in existing OOD segmentation approaches: their inability to comprehend visual scenes contextually. Current methods predominantly rely on local pixel-level uncertainty metrics or global scene statistics, rather than reasoning about object relationships, scene composition, and spatial dependencies—critical elements for understanding fine-grained semantics in complex environments.}

\yj{To address the problems}, motivated by the recent breakthroughs in Large Language Models (LLMs) such as GPT-4~\cite{achiam2023gpt} and their reasoning capability, we explore the application of Chain-of-Thought (CoT)~\cite{wei2022chain} reasoning to the relatively underexplored area of OOD semantic segmentation in road scene understanding. We introduce a novel framework that employs GPT-4-driven CoT prompting specifically tailored for GroundedSAM~\cite{ren2024grounded}, a combination of GroundingDINO~\cite{liu2024grounding} and Segment Anything Model (SAM)~\cite{kirillov2023segment}. Our proposed framework exploits the extensive knowledge and advanced logical reasoning capabilities of GPT-4~\cite{achiam2023gpt} to generate highly effective text prompts. \dw{These prompts provide critical contextual cues that effectively guide the GroundedSAM model in identifying regions of interest, thereby significantly improving both segmentation performance and interpretability.}


The primary contributions of our research are:

\begin{itemize}
    \item Clearly identifying and defining the problematic scenarios where current OOD semantic segmentation models consistently underperform, verified through comprehensive analysis of the RoadAnomaly dataset.
    \item 
    \dw{Proposing a GPT-4-based CoT prompting framework that leverages advanced prompt engineering techniques to improve OOD detection and segmentation performance.}
    
    \item \dw{The proposed framework enhances interpretability and robustness of OOD segmentation by combining CoT-guided reasoning with open-vocabulary detectors.}
\end{itemize}


\section{Related work}
\subsection{OOD Semantic Segmentation}

Out-of-distribution (OOD) semantic segmentation~\cite{zhao2024segment,tian2022pixel,liu2023residual,liang2022gmmseg,nayal2023rba,rai2023unmasking,hendrycks2019scaling,sinhamahapatra2024finding} aims to segment regions belonging to unknown classes not encountered during training, a critical task in safety-sensitive domains like autonomous driving.

Recent methods include entropy-based, score-based, hybrid, and mask-based. Entropy-based methods, such as Meta-OoD~\cite{chan2021entropy}, model prediction uncertainty through entropy maximization. Score-based methods like PEBAL~\cite{tian2022pixel} and RPL~\cite{liu2023residual} employ pixel-level anomaly scoring via energy functions or residual features. Hybrid approaches, notably DenseHybrid~\cite{grcic2022densehybrid}, integrate discriminative and generative signals, though often sensitive to threshold tuning. Mask-based methods, including RbA~\cite{nayal2023rba}, Mask2Anomaly~\cite{rai2023unmasking}, and S2M~\cite{zhao2024segment}, achieve strong object-level localization through mask classification or abstention strategies, effectively reducing false positives near ambiguous regions.

Unlike previous methods, we propose a novel inference-driven framework leveraging Chain-of-Thought (CoT) reasoning~\cite{wei2022chain} via Large Language Models (LLMs), such as GPT-4~\cite{achiam2023gpt}. Our approach guides segmentation through structured visual reasoning with semantically grounded prompts, enabling robust OOD detection in both general and challenging scenarios.

\subsection{CoT Reasoning in Vision Tasks Using LLMs}


Large Vision-Language Models (LVLMs), such as LLaVA~\cite{liu2023visual}, GPT-series~\cite{achiam2023gpt,brown2020language}, Kosmos~\cite{peng2023kosmos}, and Flamingo~\cite{alayrac2022flamingo}, have significantly advanced multimodal understanding through visual perception and high-level language reasoning. These models excel in \dw{vision-language tasks} such as visual question answering and image captioning.

To improve interpretability and localized reasoning, recent studies have extended Chain-of-Thought (CoT)~\cite{wei2022chain} prompting to vision tasks. Works such as Visual CoT~\cite{shao2024visual}, Chain-of-Spot~\cite{liu2024chain}
and CCoT~\cite{mitra2024compositional} guide the reasoning process via intermediate rationales linked to spatial regions, while Shikra~\cite{chen2023shikra} enables fine-grained coordinate-level grounding within dialogues. Inspired by these approaches, we propose a framework that leverages 
step-by-step reasoning to infer task-relevant Regions of Interest (ROIs) in complex road scenes. By explicitly directing attention to visually challenging areas, our method enhances OOD object identification and improves segmentation robustness in safety-critical scenarios.

\section{Method}

\begin{figure}[t]
    \centering
    \includegraphics[width=\linewidth]{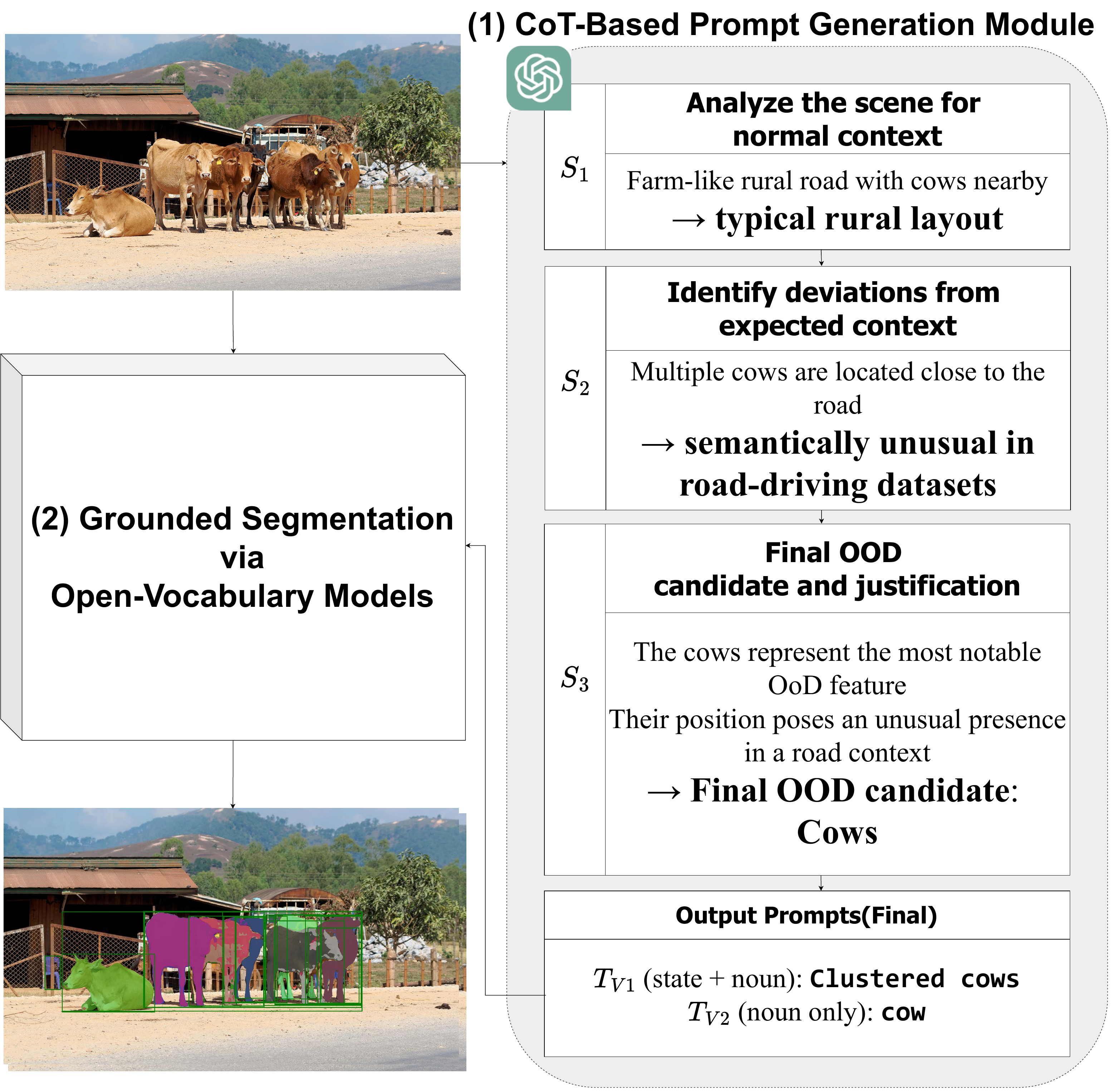}
    \caption{Overview of the proposed CoT-Segmenter framework. Given a road scene image, our CoT-Based Prompt Generation Module (right) performs step-by-step reasoning via GPT-4 to identify semantic anomalies and outputs a concise text prompt in two formats: V1 (state + noun) and V2 (noun only). Each prompt is independently fed into the Grounded Segmentation Module (left), which employs open-vocabulary detectors (GroundedSAM) to produce segmentation masks. The final OOD prediction is obtained by taking the union of the pixel regions from both prompts.}
    \label{fig:Archi}
    \vspace{-7mm}
\end{figure}

\subsection{Overview} 
We propose a novel two-stage framework for OOD object segmentation in road scenes. The core idea is to exploit the reasoning capability of a large vision-language model via Chain-of-Thought (CoT)~\cite{wei2022chain} prompting to generate text-based soft supervision, which guides a powerful segmentation backbone, GroundedSAM~\cite{ren2024grounded}. Our framework consists of two main modules:

\textbf{CoT-Based Prompt Generation Module:}  
Given an input road scene image denoted by \(I\), we use a large language model to infer OOD object candidates through step-by-step visual reasoning. This process unfolds over three structured stages, producing intermediate reasoning outputs \(S_1\), \(S_2\), and \(S_3\). Based on the final reasoning result \(S_3\), the model generates two textual prompts: \(T_{\mathrm{V1}}\), a descriptive phrase, and \(T_{\mathrm{V2}}\), concise noun label \dw{indicating the expected OOD object class.} These prompts serve as language queries for grounded segmentation module by the open-vocabulary object detectors.

\textbf{Grounded Segmentation Module (GroundedSAM):}  
Using the generated prompts \((T_{\mathrm{V1}}, T_{\mathrm{V2}})\), we apply GroundingDINO~\cite{liu2024grounding} to retrieve bounding boxes \(B\) that align with the semantics of the prompts. The predicted boxes are subsequently refined by \dw{Segment Anything Model} (SAM)~\cite{kirillov2023segment} to produce pixel-level segmentation masks. We obtain the final prediction \(M_{\mathrm{final}}\) by performing segmentation independently for \dw{both $T_{\mathrm{V1}}$ and $T_{\mathrm{V2}}$}. 
Finally, we apply the union operation to each resulting mask, respectively.
This architecture enables flexible and scalable OOD segmentation without category-specific training. It performs robustly in challenging scenarios such as densely clustered objects, small/far anomalies, and large dominant foreground objects, while maintaining strong generalization across a range of OOD segmentation benchmarks.

\subsection{CoT-Based Prompt Generation}


To enhance the performance of OOD object detection, we leverage the visual CoT
reasoning capability of GPT-4~\cite{achiam2023gpt} to generate informative and interpretable text prompts for segmentation supervision. 
\dw{Through a step-by-step CoT reasoning process, the proposed method systematically recognizes the typical elements of road scenes and identifies visually or contextually abnormal objects and regions.}
The CoT process is decomposed into three sequential stages:

\begin{equation}
\begin{aligned}
S_1 &= R_1(I), \\
S_2 &= R_2(S_1, I), \\
S_3 &= R_3(S_2, I), \\
(T_{\mathrm{V1}}, T_{\mathrm{V2}}) &= f_{\mathrm{prompt}}(S_3).
\end{aligned}
\label{eq:cot}
\end{equation}

\begin{table*}[t]
\centering
\caption{Performance degradation on RoadAnomaly under challenging scenarios. We compare OOD segmentation models on the original RoadAnomaly dataset (Standard) and our proposed challenging subset (Challenging), which is derived from RoadAnomaly and contains difficult scenarios where existing models consistently show degraded performance. All models exhibit significant performance drops under these challenging conditions, while our CoT-based framework demonstrates the most robust performance among them.}
\vspace{4mm}
\resizebox{0.85\textwidth}{!}{
\begin{tabular}{l|l|cc|cc}
\hline
\multirow{2}{*}{Method} 
& \multirow{2}{*}{Type}
& \multicolumn{2}{c|}{mIoU($\uparrow$)} 
& \multicolumn{2}{c}{F1 Score($\uparrow$)} \\
& & Standard & Challenging
  & Standard & Challenging \\
\hline
S2M~\cite{zhao2024segment}           & Mask-based       & 0.591 & 0.417 & 0.591 & 0.440 \\
RbA~\cite{nayal2023rba}              & Mask-based       & 0.413 & 0.353 & 0.584 & 0.522 \\
Mask2Anomaly~\cite{rai2023unmasking} & Mask-based       & 0.431 & 0.386 & 0.561 & 0.452 \\
SynBoost~\cite{di2021pixel}          & Generative-based & 0.272 & 0.229 & 0.293 & 0.237 \\
RPL+CoroCL~\cite{liu2023residual}    & score-based      & 0.415 & 0.360 & 0.535 & 0.473 \\
PEBAL~\cite{tian2022pixel}           & score-based      & 0.377 & 0.353 & 0.232 & 0.214 \\
\hline
GroundedSAM(`object')                & Prompt-based     & 0.803 & 0.724 & 0.862 & 0.799 \\
GroundedSAM(CoT)(ours)               & Prompt-based     & \textbf{0.912} & \textbf{0.838} 
                                                        & \textbf{0.941} & \textbf{0.892} \\
\hline
\end{tabular}
}
\label{tab:tab1}
\end{table*}

Here, \(R_1\), \(R_2\), and \(R_3\) denote reasoning functions implemented by the language model, progressively refining the understanding of the input image \(I\). Specifically, as shown in 
Figure~\ref{fig:Archi}, the first step \(R_1\) performs a comprehensive scene analysis to understand the typical layout and common semantic components to guide the detection of OOD objects in road environments. This includes identifying normal object attributes such as expected shapes, materials, and contextual relationships typically observed in driving scenes. 

The second step \(R_2\) identifies abnormal regions or objects that deviate from \dw{common semantics}. These are categorized along three axes: appearance mismatch (e.g., texture, shape, color, or material distinct from surrounding objects), semantic 
inconsistency (e.g., objects unlikely to appear in a road context), and spatial irregularity (e.g., abnormal size, position, or scale).
By integrating these criteria, the third step \(R_3\) selects visually salient and semantically inconsistent regions that are most likely to represent OOD instances.
Based on the final reasoning output \(S_3\), the model generates two complementary types of prompts: (1) \(T_{\mathrm{V1}}\), a descriptive state+noun phrase (e.g., `scattered rocks', `overlapping sheep'), and (2) \(T_{\mathrm{V2}}\), a single noun (e.g., `rock', `sheep'). 
Prior study~\cite{ilyas2024potential} have shown that long and complex prompts often lead to suboptimal responses from models like GroundingDINO~\cite{liu2024grounding}, which tend to misinterpret overly detailed phrases. 
\dw{In contrast, our dual-prompt strategy (V1 + V2) balances specificity and clarity, resulting in consistently improved detection accuracy.}

We note that our CoT-based strategy leverages the step-by-step reasoning capability of large language models to identify regions that are semantically and spatially likely to be out-of-distribution. Our framework guides the model to analyze the visual context, determine where to focus, identify what appears unusual, and understand why it may be considered OOD. As a result, the generated prompts become more focused and informative. This approach not only provides interpretable reasoning but also offers an effective mechanism for directing the detection pipeline toward OOD targets. The corresponding experimental results are presented in Table~\ref{tab:tab3}.

\subsection{Zero-shot Grounded Segmentation}

Given the CoT-generated prompts \((T_{\mathrm{V1}}, T_{\mathrm{V2}})\), we utilize the GroundedSAM~\cite{ren2024grounded} pipeline, which integrates GroundingDINO~\cite{liu2024grounding} and SAM~\cite{kirillov2023segment}, to obtain pixel-level OOD segmentations. Specifically, GroundingDINO retrieves bounding boxes \(\mathcal{B}\) from the image \(I\) and text prompts \(T_{\mathrm{V_i}} \in \{T_{\mathrm{V1}}, T_{\mathrm{V2}}\}\) as:
\begin{equation}
\mathcal{{B\mathrm{_i}}} = G(I, T_{\mathrm{V_i}}),
\end{equation}
where \(G(\cdot)\) denotes the GroundingDINO model. These predicted boxes are then passed to SAM to generate fine-grained segmentation masks \(\mathcal{M}\), defined as:
\begin{equation}
\mathcal{M_{\mathrm{V_i}}} = S(I, {B\mathrm{_i}}),
\end{equation}
where \(S(\cdot)\) denotes the SAM module. This two-stage framework enables OOD segmentation in a zero-shot manner, guided by CoT-informed textual reasoning.


This architecture enables interpretable and category-agnostic segmentation in a zero-shot setting. By leveraging the intermediate bounding box representation, it facilitates alignment between visual regions and natural language prompts, which is particularly beneficial for identifying ambiguous or spatially irregular OOD objects.
In practice, we apply both descriptive \((T_{\mathrm{V1}})\) and concise \((T_{\mathrm{V2}})\) prompts generated by \dw{the CoT-based prompt generation module}
and take the union of the resulting segmentation masks to construct the final prediction:

\begin{equation}
M_{\mathrm{final}} = M_{\mathrm{V1}} \cup M_{\mathrm{V2}},
\end{equation}

This dual-prompt strategy ensures broader coverage and improved robustness by leveraging the complementary nature of the prompts.





\begin{figure*}[t]
    \centering
    \includegraphics[width=\textwidth]{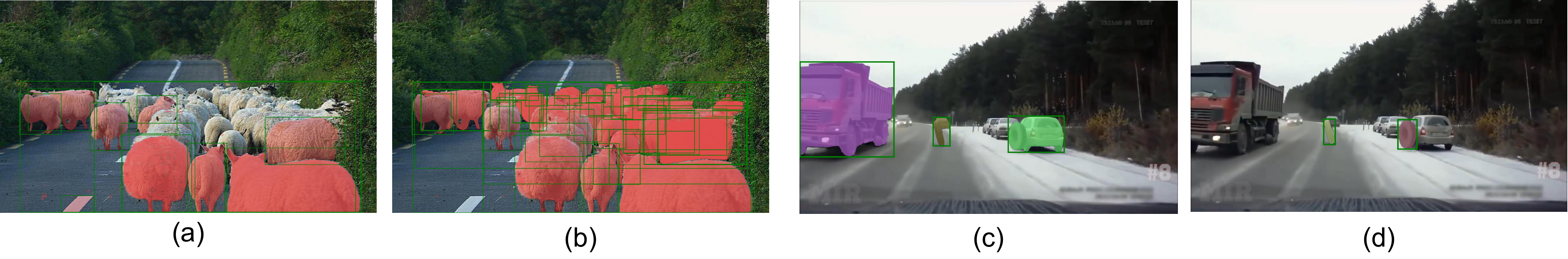}
    \caption{Comparison of segmentation results using different prompts: (a), (c) use a generic prompt (`object’), while (b), (d) use CoT-derived prompts. Prompts in (b) and (d) are V1: ``Dense sheep blocking”, V2: ``sheep” and V1: ``Uncontrolled rolling tires”, V2: ``tire”, respectively.}
    \label{fig:input_prompt}
\end{figure*}

\section{Experiment}

\subsection{Experimental Setup}\label{sec:experiment_setup}
We conduct experiments on the RoadAnomaly dataset~\cite{lis2019detecting} and a curated subset derived from it, which specifically includes scenes containing (1) densely overlapping objects, (2) small and distant anomalies, and (3) large foreground-dominant instances. We refer to this subset as the Challenging dataset, as it is designed to evaluate performance under scenarios where existing OOD segmentation methods consistently struggle.
For evaluation, we report mean Intersection over Union (mIoU) and F1 score, following standard protocol in OOD segmentation.
Our framework employs GroundedSAM~\cite{ren2024grounded}, which integrates GroundingDINO~\cite{liu2024grounding} with a Swin-T backbone for open-vocabulary object detection, and SAM~\cite{kirillov2023segment} with the ViT-H variant for fine-grained segmentation. CoT reasoning is performed via GPT-4~\cite{achiam2023gpt}, generating both V1 (state+noun) and V2 (noun-only) prompts. For each image, the box threshold and text threshold values in GroundedSAM were individually optimized to maximize segmentation accuracy, reflecting the sensitivity of open-vocabulary grounding to threshold tuning.
We perform the experiments using a single NVIDIA GeForce RTX 4070 GPU. 

\subsection{Performance under Challenging Scenarios}

In this section, we mainly investigate whether existing OOD segmentation models consistently exhibit performance degradation under challenging conditions. Specifically, we focus on three problematic scenarios outlined in Section.~\ref{sec:experiment_setup}.
As shown in Table~\ref{tab:tab1}, our experimental results demonstrate that all evaluated OOD segmentation models commonly suffer performance degradation under these conditions. Our proposed CoT-based framework achieves the highest segmentation accuracy among all evaluated approaches, confirming its effectiveness in addressing these challenging conditions.

\subsection{Comparative Analysis on Text Prompt}

\begin{table}[h]
\centering
\vspace{-2mm}
\caption{Segmentation performance comparison across different types of text prompts.}
\vspace{2mm}
\resizebox{1\linewidth}{!}{
\begin{tabular}{l|cc}
\hline
Text Prompt                                   & mIoU & F1 score \\
\hline
`Object'                                      & 0.724 & 0.799   \\
`OOD object . object'                         & 0.759 & 0.828   \\
`animal . cone . rock . object'               & 0.770 & 0.826   \\
`animal . cone . rock . object . OOD object'  & 0.785 & 0.837   \\
`OOD object'                                  & 0.790 & 0.857   \\
\hline
only V1                                       & 0.787 & 0.858   \\
only V2                                       & 0.777 & 0.831   \\
\hline
\( V1 \cup V2 \) (Ours)                       & \textbf{0.838} & \textbf{0.892} \\
\hline
\end{tabular}
}
\label{tab:tab2}
\end{table}

\begin{table}[h]
\centering
\vspace{-4mm}
\caption{Ablation study of CoT-based prompt generation steps.}
\vspace{2mm}
\begin{tabular}{l|cc}
\hline
Text Prompt         & mIoU & F1 score \\
\hline
Non-CoT(`object')   & 0.724 & 0.799  \\
\hline
only GPT-4           & 0.380 & 0.473  \\
\hline
1 step              & 0.088 & 0.146  \\
1,2 step            & 0.483 & 0.580 \\
\hline
1,2,3 step          & \textbf{0.838} & \textbf{0.892} \\
\hline
\end{tabular}
\label{tab:tab3}
\end{table}  

In this subsection, we investigate how the choice of text prompts influences the performance of the open-vocabulary object detection model, GroundedSAM~\cite{ren2024grounded}. To comprehensively analyze this impact, we conducted comparative analyses using several prompt types: a single general noun prompt (`object'), a more descriptive ensemble prompt (e.g., `animal . cone . rock . object'). As shown in Table~\ref{tab:tab2}, performance progressively improved from using the simple noun prompt to the ensemble prompts, achieving the highest scores when explicitly mentioning the OOD context.

Furthermore, we conducted an ablation study evaluating the individual performance of V1-style (state + noun) prompts and V2-style (single noun) prompts, as well as their combined effect. Table~\ref{tab:tab2} demonstrates that the combination of V1 and V2 prompts ($V1 \cup V2$) yielded the best results, indicating complementary strengths in leveraging detailed context and simplicity.

We additionally explored the effectiveness of CoT-based prompt generation, as shown in Table~\ref{tab:tab3}. The Non-CoT approach uses a generic prompt (`object'), serving as a baseline. The only GPT-4~\cite{achiam2023gpt} condition involves directly querying GPT-4 to identify the OOD object without structured reasoning. Subsequent experiments progressively integrated intermediate reasoning steps from our CoT-based prompt generation module. Notably, the complete CoT pipeline, incorporating all intermediate reasoning steps (steps 1-3), consistently achieved the highest segmentation accuracy, demonstrating the value of structured, step-by-step visual reasoning.

Qualitatively, Figure~\ref{fig:input_prompt} provides a comparison of segmentation results using generic prompts versus CoT-derived prompts. Images processed with CoT-generated prompts exhibit clearer and more precise segmentation, highlighting the practical benefits of detailed and context-aware prompt generation strategies in challenging road scenarios.

\section{Conclusion}

We proposed a CoT-based framework that leverages the reasoning ability of LLMs to enhance OOD segmentation in complex road scenes. By generating structured prompts based on visual understanding, our method effectively guides the text-based open-vocabulary segmentation model, showing improved performance, especially in known challenging scenarios.
Despite these improvements, a notable limitation of our approach lies in the use of overly detailed or lengthy prompts, which can degrade detection performance. This is primarily because the model attempts to localize every word in the prompt, often resulting in oversegmentation or misaligned bounding boxes.
Future work will investigate strategies to better align language-driven reasoning with grounding behavior, such as selectively filtering prompt content or integrating intermediate reasoning steps prior to grounding.


\section*{Acknowledgment}
This work was supported by the Institute of Information $\&$ Communications Technology Planning $\&$ Evaluation (IITP) grant funded by the Korea government (MSIT) [RS-2021-II211341, Artificial Intelligent Graduate School Program (Chung-Ang University) and RS-2022-II220124, Development of Artificial Intelligence Technology for Self-Improving Competency-Aware Learning Capabilities].

\bibliographystyle{ieee}
\bibliography{egbib}
\end{document}